\documentclass[conference]{IEEEtran}
\IEEEoverridecommandlockouts
\usepackage{amsmath,amssymb,amsfonts}
\usepackage{algorithmic}
\usepackage{graphicx}
\usepackage{textcomp}

\usepackage[table,xcdraw]{xcolor}

\usepackage{todonotes}
\usepackage{multirow}
\usepackage{changepage}   
\usepackage[numbers,sort]{natbib}

\def\BibTeX{{\rm B\kern-.05em{\sc i\kern-.025em b}\kern-.08em
    T\kern-.1667em\lower.7ex\hbox{E}\kern-.125emX}}
\begin{document}

\title{The Weak Supervision Landscape\\
\thanks{Partially supported by project TEC2017-83838-R, funded by FEDER /Ministerio de Ciencia, Innovación y Universidades – AEI;
by the SPHERE Next Steps Project funded by EPSRC [grant EP/R005273/1]; RP and RSR are funded by
the UKRI Turing AI Fellowship EP/V024817/1.
}}


 \author{\IEEEauthorblockN{Rafael Poyiadzi\textsuperscript{1}, Daniel Bacaicoa-Barber\textsuperscript{2}, Jesus Cid-Sueiro\textsuperscript{2},\\
 Miquel Perello-Nieto\textsuperscript{1}, Peter Flach\textsuperscript{1}, Raul Santos-Rodriguez\textsuperscript{1}}
 \IEEEauthorblockA{\textit{\textsuperscript{1}Intelligent Systems Lab}, \textit{University of Bristol}, Bristol, UK \\
 \textit{\textsuperscript{2}Signal Theory and Communications Dept.}, \textit{Universidad Carlos III of Madrid}, Spain\\
 Contact Email: rp13102@bristol.ac.uk
}
}

\maketitle

\begin{abstract}
Many ways of annotating a dataset for machine learning classification tasks that go beyond the usual class labels exist in practice. These are of interest as they can simplify or facilitate the collection of annotations, while not greatly affecting the resulting machine learning model. Many of these fall under the umbrella term of weak labels or annotations. However, it is not always clear how different alternatives are related. In this paper we propose a framework for categorising weak supervision settings with the aim of: (1) helping the dataset owner or annotator navigate through the available options within weak supervision when prescribing an annotation process, and (2) describing existing annotations for a dataset to machine learning practitioners so that we allow them to understand the implications for the learning process. To this end, we identify the key elements that characterise weak supervision and devise a series of dimensions that categorise most of the existing approaches. We show how common settings in the literature fit within the framework and discuss its possible uses in practice.  
\end{abstract}


\section{Introduction}



A machine learning classification task requires having a dataset of pairs of instances and labels. Obtaining labels of good quality can be expensive, time-consuming, or difficult in general. These constraints have led to the development and study of several flexible settings where annotations are assumed to not be perfect, but still suitable for the learning process. These are usually referred to as weak supervision (WS). For example, specialised products like Amazon's Mechanical Turk  provide access to pools of (non-expert) annotators whose labels are sometimes ambiguous and noisy. Interestingly, weak labels can take several forms and not only be the result of the annotation process (e.g., the data not annotated by an expert, or automatically extracted from the web) but they can come from the choices made by the dataset owner when deciding on the annotation process (e.g., allow annotators to provide more than one candidate classes when uncertain). 


In machine learning, weakly supervised classification refers the task of obtaining a classifier from a given \textit{weak} dataset, such that it has a low generalisation error with respect to the true data distribution. Even though weakly supervised learning encompasses settings which are widely applicable and studied, it is yet to become a standard machine learning setting such as traditional supervised classification or clustering \cite{patrini2016weakly}. They have been previously referred to as indirect supervision \cite{raghunathan2016estimation}, distant supervision \cite{surdeanu2012multi}, inaccurate, incomplete, or inexact supervision \cite{zhou2018brief}, learning from measurements \cite{liang2009learning}, learning imprecise and fuzzy observations \cite{hullermeier2014learning} and many more. In this work we refer to all the approaches where the observed label is not perfect as \texttt{weak}. Where applicable we will also refer to the observed label as \texttt{weak}, as opposed to the unobserved \texttt{clean label}. We will also be referring to the process by which a clean label is transformed to a weak label as the \texttt{weakening process}. 

In this paper we introduce a framework for categorising weak supervision settings by identifying the key set of dimensions that should be used to describe the different instantiations that exist in the literature and in practice. This framework can then be used by both dataset owners / annotators and machine learning practitioners / researchers. For the former, it will be a tool to navigate the landscape of options when designing the data collection or annotation process or to describe an existing dataset. In both cases it will aid communication and can help future users understand implications of the type of annotations present in the dataset. For the later, it can be used to find the right place for a new or existing technique to learn from weak labels and identify (open) research problems in the field and have a clearer understanding of the generalisation of their contributions.

The remainder of the paper is structured as follows: in Section \ref{section:dimensions} we introduce the framework and in Section \ref{section:examples} we present several well studied WSL settings and discuss how they fit within our framework. Lastly in Section \ref{section:conclusion} we discuss the main implications and limitations of the work and directions for future research.

\begin{table*}[]
\renewcommand{\arraystretch}{1.35}
\centering
\caption{Framework for categorising weak supervision settings constituted by categories and dimensions. The third column lists options for each. Individual settings can be identified by selecting one option per dimension. We note that certain combinations of options are not compatible. The final column summarizes the meaning of each of the dimensions for use in the prescription or description of the corresponding dimension.}
\vspace{-4pt}
\begin{tabular}{|c|c|c|p{8cm}|}
\hline
\rowcolor[HTML]{EFEFEF} 
Category                            & Dimension                                                              & Options              & \multicolumn{1}{c|}{\cellcolor[HTML]{EFEFEF}Question}                                                                                        \\ \hline\hline
                                    & \begin{tabular}[c]{@{}c@{}}Number of \\ classes\end{tabular}           & Binary / Multi-class & How many classes does the task involve?                                                                                                         \\ \cline{2-4} 
\multirow{-3}{*}{True Label Space}  & Multi-label                                                            & Yes / No             & Can instances belong to more than one class?                                                                                                 \\ \hline\hline
                                    & Unsupervised                                                           & Yes / No             & \begin{tabular}[c]{@{}l@{}}Are annotators allowed to not annotate certain samples?\end{tabular} \\ \cline{2-4} 
                                    & Soft labels                                                           & Yes / No             & \begin{tabular}[c]{@{}l@{}}Are annotators allowed to use soft or probabilistic labels?\end{tabular} \\ \cline{2-4} 
                                    & \begin{tabular}[c]{@{}c@{}}Number of\\ Annotators\end{tabular}          & $1~/~\textgreater1$              & How many annotators will annotate the data?                                                                                            \\ \cline{2-4} 
\multirow{-5}{*}{Weak Label space}  & \begin{tabular}[c]{@{}c@{}}Number of \\ candidate classes\end{tabular} & $1~/~\textgreater1$                 & Are annotators allowed to provide annotations covering more than one class?                                                                 \\ \hline\hline
                                    & Aggregation                                                    & Yes / No             & Are samples annotated individually or as a group?                                                                                                              \\ \cline{2-4} 
                                    & Class dependent                                                        & Yes / No             & Are classes equally prone to annotation errors?                                                                                                                                             \\ \cline{2-4} 
\multirow{-3}{*}{Weakening Process} & Instance dependent                                                             & Yes / No             & Are samples equally prone to annotation errors?                                                                                                                        \\ \hline
\end{tabular}
\label{table:dimensions}
\renewcommand{\arraystretch}{1}
\end{table*}










\section{Dimensions of Weak Supervision}
\label{section:dimensions}

In this section we present the dimensions of weak supervision, which are the building blocks of our framework. We separate the dimensions into three groups based on whether they refer to the true label space, the weak label space or the weakening process. Table~\ref{table:dimensions} condenses all this information together with questions to illustrate the meaning of each dimension and the options that they offer. This table can be used as a standalone tool to plan or understand weak annotations.

\subsection{True Label Space}
\label{section:true_label_space}
Here we present dimensions that derive directly form the nature and description of the task. Classification tasks can take different forms, from binary classification, where we might be interested in identifying an image as a dog or a cat, to multi-class, where we might aim to recognize images of dogs according to their breed. This is summarized in the first dimension -- \textbf{number of classes}. Additionally, we accept that it is sometimes the case that more than one categories can be assigned to a single instance as in \textbf{multi-label} settings, which forms the second dimension. The types of true/clean label that we consider are then as follows.
\begin{align}
    &\textrm{Number of classes:}\nonumber\\
    &\hspace{15pt}\mathcal{Y}_k~=~ \big\{\boldsymbol{y}~\mid~\boldsymbol{y}\in\{0,~1\}^k,~\boldsymbol{1}^\top \boldsymbol{y}=1\big\}\label{eq:multiclass_label_space}\\
    &\textrm{Multi-label:}\nonumber\\
    &\hspace{15pt}\mathcal{Y}_{m, k} ~=~ \big\{\boldsymbol{y}~|~\boldsymbol{y}\in\{0,~1\}^k,~1\leq\boldsymbol{1}^{\top}\boldsymbol{y}\leq m\leq k\big\}\label{eq:multilabel_label_space}
\end{align}
Although this can be extended to structured data, we do not explore it here for simplicity. We also assume that the dataset only contains what is defined by the task, e.g., in the case of the task being classifying dogs vs cats, there would not be images of other animals in the dataset. Even though we do not discuss the learning stage in this paper, the framework is constructed  keeping in mind that part of the task would be to obtain a classifier: $f:~\mathcal{X}~\to~\mathcal{Y}_{\textrm{Clean}}$ and in minimizing ${\mathbb{E}_{X,Y}\ell(Y, f(X))}$, where $\ell(.,.)$ is a loss function and the expectation is over the clean label data distribution.

\subsection{Weak Label Space}
\label{section:weak_label_space}

In this section we focus on dimensions concerned with the weak label space, $\mathcal{Y}_{Weak}$ (See Eq.~\ref{eq:noise_process_2}) and the possible forms that it might take. These characterise the degrees of freedom of the annotator.

\noindent\textit{Access to unlabelled data.} In certain situations, besides the potentially weakly labelled dataset, we also have access to a separate unlabelled dataset. This could be either because annotators are allowed to return an empty annotation, or because this set of data was just not chosen for annotation. We refer to this dimension as \textbf{unsupervised}.

\noindent\textit{Access to multiple annotators.}
We usually assume a dataset is annotated by one annotator, but it might be the case that a \textbf{number of annotators} provide annotations for the same dataset. Annotations on the same instance will not necessarily agree, potentially creating ambiguity.

\noindent\textit{Restriction on number of assigned candidate classes.} In binary and multi-class classification, the classes are mutually exclusive and exhaustive, which means that every instance is associated with one true class only. In the weak label space, we could allow an annotator to provide a set of candidate classes, instead of just one (\textbf{number of candidate classes}).

\begin{figure}[!t]
    \centering
    \includegraphics[width=\linewidth]{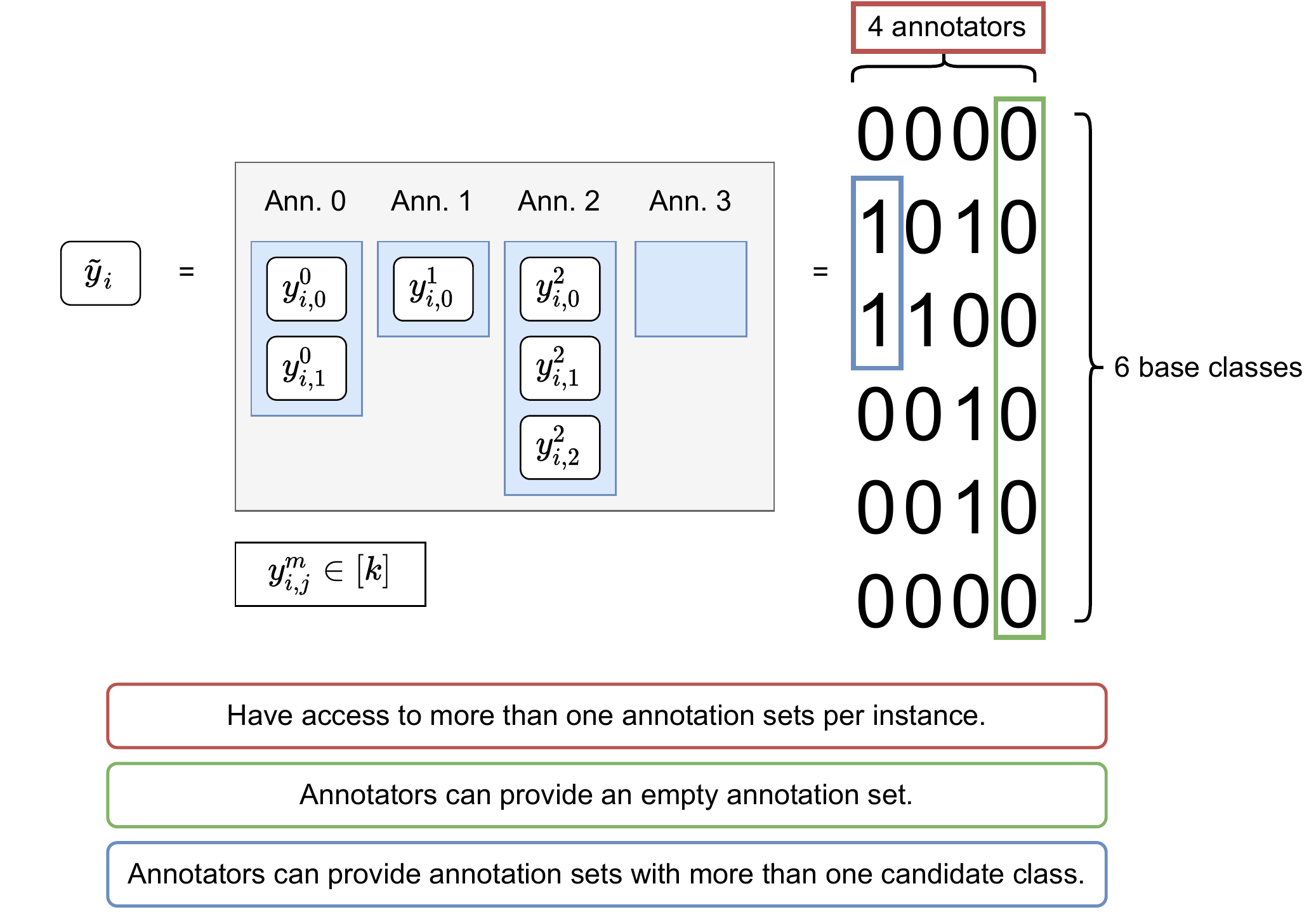}
    \caption{An example of a non-aggregate weak label. For this case we consider six base classes and access to four annotators. This example showcases three dimensions: (1) having access to more than one annotators, (2) annotators can provide an empty annotation set (see annotator 3), and (3) annotators can provide an annotation set with more than one candidate class (see annotators 1 and 2).}
    \label{fig:weak_label_example}
\end{figure}

An example of a weak label and how this set of dimensions can affect what we get to observe is shown in Figure \ref{fig:weak_label_example}. It shows how these three dimensions can increase the complexity of a weak label, as we can have several annotators, allow them to provide no annotation, and also allow them to assign more than one class to each sample. 

\noindent\textit{Soft labels.} Also known as probabilistic labels, the \textbf{soft labels} relaxation allows for more flexibility in the annotation process by letting an annotator express a degree of belief. For example, instead of resorting to `dog' for an image, they could say "70\% confident this image contains a dog" \cite{peng2014learning}.

\subsection{Weakening Process}
\label{section:noise_function_properties}

The final set of dimensions captures the different aspects of the weakening process, i.e., the (usually unknown) transformation the maps true/clean to weak labels. Interestingly, the weakening process depends on the annotator and how they make suboptimal annotation decisions, but it can also depend on choices made by the dataset owner or on the task itself. 

\begin{figure}[!t]
    \centering
    \includegraphics[width=0.85\linewidth]{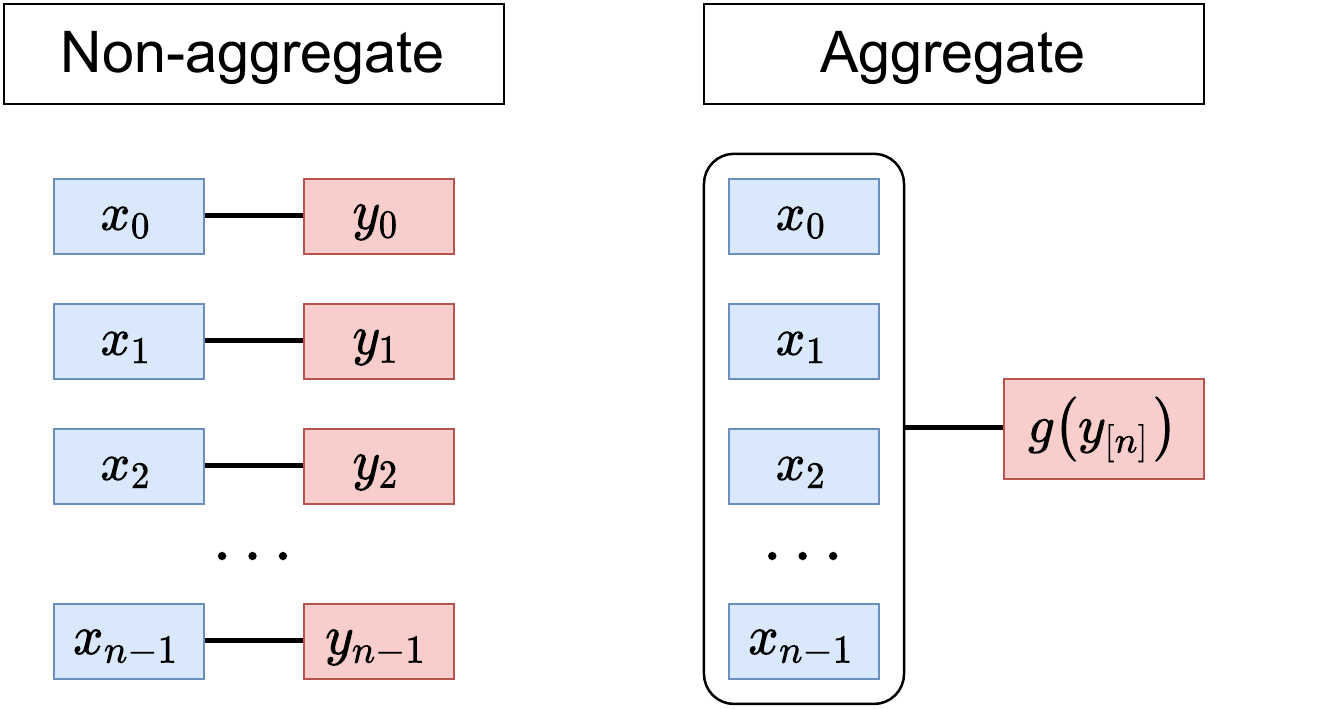}
    \caption{Depiction of the difference between non-aggregate and aggregate settings. In the case of non-aggregate settings, there is a one-to-one relationship between instances and labels, while in the case of aggregate noise settings we observe one label per group. We note that a label could be one class, multiple or none, as we describe below.}
     \label{fig:agg_vs_nonagg}
\end{figure}

\noindent\textit{Aggregation.} 
A key dimension is whether the labels provided correspond to a single instance, or whether they depend on, and correspond to, a set of instances through an \textbf{aggregation} mechanism. For the aggregate cases, the labels provided correspond to a set of instances which we will refer to as a `bag'. Instead of being provided with a set of instance-label pairs $\{(x_i,~y_i)\}_{i=1}^n$, we are provided with a set of $\{(x_i,~b_i)\}_{i=1}^n$ where $b_i$'s are bag indicators and  $\{(b_j,~t_j)\}_{j=1}^m$, where $t_j$'s are the corresponding labels. The labels in this case are of the form:
\begin{equation}
    \label{eq:aggregate_function}
    t_j~=~g\Big(\big\{y_i~\mid~i\in b_j\big\}\Big)
\end{equation}
where we use $i\in b_j$ to imply that the $i^{th}$ instance belongs to the $j^{th}$ bag, and $g$ is the label aggregation function. It should be noted that $t_j$ is the label for \textit{all} instances $x_i,~i\in b_j$. 
The distinction between the two is shown pictorially in Fig.~\ref{fig:agg_vs_nonagg}. 

For non-aggregate settings, we will refer to $\mathbb{P}(\tilde{Y}\mid X,Y)$, where $\tilde{Y}$ denotes the random variable for the weak label, as the \textit{weakening function}, which upon conditioning has the form,
\begin{equation}
    \label{eq:noise_process}
    \tau(y;x)=\mathbb{P}\big(\tilde{Y}\mid X=x,Y=y\big)
\end{equation}
with
\begin{equation}
    \label{eq:noise_process_2}
    \tau(y;x): \mathcal{Y}_{Clean}\to\triangle_{\mathcal{Y}_{Weak}}
\end{equation}
where $\triangle_{\mathcal{Y}_{Weak}}$ is the probability simplex over $\mathcal{Y}_{Weak}$. Eqs. \ref{eq:noise_process} \& \ref{eq:noise_process_2} involve three objects: 
\begin{itemize}
    \item the clean label space $\mathcal{Y}_{Clean}$ (covered in Sec.~\ref{section:true_label_space}),
    \item the weak label space $\mathcal{Y}_{Weak}$ (covered in Sec.~\ref{section:weak_label_space}) and
    \item the weakening function $\tau(y;x)$.
\end{itemize}
The weakening function acts between discrete sets and can therefore be described as a \textit{Categorical} (\texttt{Cat}) distribution. This distribution is parametrized by a column of a \textit{mixing matrix} or transition matrix, $\boldsymbol{T}$, which is a non-negative column-wise stochastic matrix. The weakening process is modelled as,
\begin{equation}
    \label{eq:noise_categorigal}
    \tilde{Y}~\mid~Y=\boldsymbol{y},~X=\boldsymbol{x}~\sim~\texttt{Cat}\big(\boldsymbol{T}\boldsymbol{y}\big)
\end{equation}
With regards to dimensions arising of the weakening function we will consider two, the dependence on the instance and on the true class. 



\noindent\textit{Dependence on instance.} We consider two cases, \textbf{instance dependence} or {instance independence}~\cite{menon2018learning}.
\begin{align*}
    &\textrm{Instance-independent noise (IIN)}:\\
    &\hspace{10pt} \mathbb{P}(\tilde{Y}=\tilde{y}~\mid~X=x,~Y=y)~=~\mathbb{P}(\tilde{Y}=\tilde{y}~\mid~Y=y),\\
    &\textrm{Instance-dependent noise (IDN)}:\\
    &\hspace{10pt} \mathbb{P}(\tilde{Y}=\tilde{y}~\mid~X=x,~Y=y)~\ne~\mathbb{P}(\tilde{Y}=\tilde{y}~\mid~Y=y).
\end{align*}



\noindent\textit{Dependence on true class.} With regards to the \textbf{class dependence} we consider \textit{symmetric} (uniform) or \textit{asymmetric} (class-conditional) with respect to the original classes \cite{menon2016learning}. In the case of multi-class classification, symmetric noise would imply:
\begin{align}
    \label{eq:multiclass_cnn}
    &\mathbb{P}(\tilde{Y}=e_u\mid X=x,Y=e_i)=\mathbb{P}(\tilde{Y}=e_v\mid X=x,Y=e_i)\\
    &~~\forall u,v,i \in [k],~u\neq v\neq i. \nonumber
\end{align}
We have asymmetric label noise when this does not hold. 


\section{The Dimensions in Practice}
\label{section:examples}

We now present several well-studied settings within the field of weak supervision and discuss how they fit in the proposed framework. We first start by exploring examples that belong to the the non-aggregate category and then move to aggregate, as this separation simplifies the formulation of the settings.
\begin{table}[t!]
\centering
\caption{Non-aggregate WSL settings according to the label-space of true labels and weak labels, and the mixing matrix used to model the weakening process.}
\vspace{-4pt}
\renewcommand{\arraystretch}{1.65}
\setlength\tabcolsep{3pt}
\begin{tabular}{|c||c|c|c|}\hline
\cellcolor[HTML]{EFEFEF}\textbf{Name} & \cellcolor[HTML]{EFEFEF}\textbf{\begin{tabular}[c]{@{}c@{}}True\\ Label Space\end{tabular}} & \cellcolor[HTML]{EFEFEF}\textbf{\begin{tabular}[c]{@{}c@{}}Weak\\ Label Space\end{tabular}} & \cellcolor[HTML]{EFEFEF}\textbf{Mixing matrix}\\\hline\hline
Noisy Labels & $\mathcal{Y}_k$ & $\mathcal{Y}_k$ & $\boldsymbol{T}\in\mathbb{R}^{k\times k}$ \\\hline
Partial Labels & $\mathcal{Y}_k$ & $\mathcal{Y}_{m, k}$ & $\boldsymbol{T}\in\mathbb{R}^{(2^k-2)\times k}$ \\\hline
Superset Learning & $\mathcal{Y}_k$ & $\mathcal{Y}_{m, k}(z)$ & $\boldsymbol{T}\in\mathbb{R}^{(2^k-2)\times k} $ \\\hline
\begin{tabular}[c]{@{}c@{}}Semi-supervised\\ Learning\end{tabular} & $\mathcal{Y}_k$ & $\mathcal{Y}_{k+1}$ & $\boldsymbol{T}\in\mathbb{R}^{(k+1)\times k}$ \\\hline
Positive-Unlabelled & $\mathcal{Y}_2$ & $\mathcal{Y}_3$ & $\boldsymbol{T}\in\mathbb{R}^{3\times 2}$ \\\hline
\begin{tabular}[c]{@{}c@{}}Multiple\\ Annotators\end{tabular} & $\mathcal{Y}_k$ & $\mathcal{Y}_k^n$ & $\Big\{\boldsymbol{T}\in\mathbb{R}^{k\times k}\Big\}_{i=1}^n$ \\\hline
\end{tabular}
\label{table:noise_settings}
\renewcommand{\arraystretch}{1}
\end{table}

\subsection{Non-aggregate WS Settings}
For non-aggregate WS settings, the selected examples are summarized in Table~\ref{table:noise_settings} to illustrate their constraints in the true label space, weak label space and weakening process. 
\paragraph{Noisy labels (Flipping noise)} In this setting, instances switch labels with a certain probability. This type of noise could be introduced when the data is labelled by a non-expert annotator, and nicely extends to having multiple such annotators. In this setting the clean and noisy label spaces are the same:
\begin{align*}
    \mathcal{Y}_k&~\to~\mathcal{Y}_k\\
    \boldsymbol{T}&~\in~\mathbb{R}^{k\times k}
\end{align*}
An example of a mixing matrix for binary classification:
\begin{equation*}
    \centering
    \bordermatrix{& \textcolor{blue}{\textrm{+}} & \textcolor{blue}{\textrm{-}}\cr
      \textcolor{blue}{\textrm{+}} & 1-\gamma_0 & \gamma_1 \cr
      \textcolor{blue}{\textrm{-}} & \gamma_0 & 1-\gamma_1}
\end{equation*}
where if $\gamma_0=\gamma_1$ it implies symmetric label noise, and asymmetric label noise otherwise. In the case of instance-independent noise, $\gamma_0$ and $\gamma_1$ are constants, i.e., all instances have their label flipped with same probability. On the other hand, with instance-dependent noise, $\gamma_0$ and $\gamma_1$ could be functions of $x$, where we could have rates differ $\gamma_0(x_i)\neq\gamma_1(x_j)$ for $x_i\ne x_j$. A comprehensive review of classification in the presence of label noise is given by \cite{frenay2013classification}.

\paragraph{Superset Learning} In this setting a weak label could be any combination of class labels, subject to the constraint that this combination contains the true label. For example, in the case of multi-class classification with three classes, for a true label $[001]$ (class 3 in one-hot encoding), we could observe $[011]$ or $[101]$, but not $[110]$ (represented as the binary OR operator on the true one-hot encoding). Superset learning has been studied in the literature under different names such as `learning with partial-labels' (see below), `learning with ambiguous labels' and `learning from complementary labels' \cite{ishida2017learning}. Some of the earlier works on the topic include \cite{grandvalet2002logistic} and \cite{jin2002learning} where the setting is referred to as `partial-labels' and `multiple labels' respectively. An example of a mixing matrix:
\begin{equation*}
    \bordermatrix{
    & \textcolor{blue}{\textrm{001}} & \textcolor{blue}{\textrm{010}} & \textcolor{blue}{\textrm{100}}\cr
  \textcolor{blue}{\textrm{001}} & \alpha_0 & 0 & 0\cr                
  \textcolor{blue}{\textrm{010}} & 0 & \beta_0 & 0\cr               
  \textcolor{blue}{\textrm{100}} & 0 & 0 & \gamma_0\cr                
  \textcolor{blue}{\textrm{110}} & 0 & \beta_1 & \gamma_1\cr         
  \textcolor{blue}{\textrm{101}} & \alpha_1 & 0 & 1-\gamma_0-\gamma_1\cr        
  \textcolor{blue}{\textrm{011}} & 1-\alpha_0-\alpha_1 & 1-\beta_0-\beta_1 & 0\cr         
 }
\end{equation*}



\begin{align*}
    \mathcal{Y}_k&~\to~\mathcal{Y}_{m,k}(z)\\
    \boldsymbol{T}&~\in~\mathbb{R}^{(2^k-2)\times k}
\end{align*}
where,
\begin{equation}
    \label{eq:superset_label_space}
    \mathcal{Y}_{m, k}(z)= \big\{\boldsymbol{y}|\boldsymbol{y}\in\{0,1\}^k,~1\leq\boldsymbol{1}^{\top}\boldsymbol{y}\leq m\leq k,~z\in \boldsymbol{y}\big\}
\end{equation}
We abuse notation and use $\boldsymbol{y}$ both as a vector and as a set and with $z\in\boldsymbol{y}$ we imply that $\boldsymbol{y}$ covers $z$, i.e. has a non-zero entry wherever $z$ has a non-zero entry.


\paragraph{Partial Labels (PLL)} PLL is sometimes used to refer to superset learning. This setting is similar to that of superset learning but where there is no restriction that the observed weak label includes the true label\cite{cid2012proper,cid2014prop}.
\begin{align*}
    \mathcal{Y}_k&~\to~\mathcal{Y}_{m,k}\\
    \boldsymbol{T}&~\in~\mathbb{R}^{(2^k-2)\times k}
\end{align*}
where,
\begin{equation}
    \label{eq:partial_label_space}
    \mathcal{Y}_{m, k} ~=~ \big\{\boldsymbol{y}~|~\boldsymbol{y}\in\{0,1\}^k,~1\leq\boldsymbol{1}^{\top}\boldsymbol{y}\leq m\leq k\big\}
\end{equation}
In both settings, the set of potential observations extends from $k$ to $2^k-2$. While standard presentations of the settings consider $2^k-1$, we choose to exclude the all inclusive potential observation and instead consider it as an extra dimension.

\paragraph{Semi-supervised Learning (SSL)} In semi-supervised learning \cite{chapelle2009semi}, on top of the usual supervised dataset, we are also provided with an unlabelled dataset. 
\begin{align*}
    \mathcal{Y}_k&~\to~\mathcal{Y}_{k+}\\
    \boldsymbol{T}&~\in~\mathbb{R}^{(k+1)\times k}
\end{align*}
where,
\begin{equation}
    \label{eq:ssl_labelspace}
    \mathcal{Y}_{k+}~=~ \big\{\boldsymbol{y}~\mid~\boldsymbol{y}\in\{0,1\}^k,~\boldsymbol{1}^\top \boldsymbol{y}\in\{0,1\}\big\}
\end{equation}
and is an extension of the multi-class label space (Eq. \ref{eq:multiclass_label_space}) that allows for no classes to be provided. An example of a mixing matrix for binary classification:

\begin{equation}
    \label{eq:ssl_mixing_matrix}
     \bordermatrix{& \textcolor{blue}{\textrm{+}} & \textcolor{blue}{\textrm{-}}\cr
        \textcolor{blue}{\textrm{+}} & 1-\gamma_0 & 0 \cr
        \textcolor{blue}{\textrm{-}} & 0 & 1-\gamma_1 \cr
        \textcolor{blue}{\textrm{na}} & \gamma_0 & \gamma_1
        }
\end{equation}

\paragraph{Positive-Unlabelled (PU) Learning}
Learning with positive and unlabelled instances \cite{liu2003} is the setting of binary classification where the training dataset only consists of positive and unlabelled instances. Situations where PU learning arises include medical records where only known previous diseases are listed and personalised advertising where visited pages and clicks are the positive cases \cite{bekker2020learning}. 
\begin{align*}
    \mathcal{Y}_2&~\to~\mathcal{Y}_{2+}\\
    \boldsymbol{T}&~\in~\mathbb{R}^{3\times 2}
\end{align*}
An example of a mixing matrix for PU learning: 
\begin{equation*}
    \bordermatrix{& \textcolor{blue}{\textrm{+}} & \textcolor{blue}{\textrm{-}}\cr
      \textcolor{blue}{\textrm{+}} & 1-\gamma_0 & 0 \cr
      \textcolor{blue}{\textrm{-}} & 0 & 0 \cr
      \textcolor{blue}{\textrm{na}} & \gamma_0 & 1}
\end{equation*}
PU learning can be seen as a special case of semi-supervised learning where $\gamma_1 = 1$ (Eq.~\ref{eq:ssl_mixing_matrix}). 
It has been extended to the multi-class case, under the name Multi-Positive and Unlabeled learning, where labeled data from multiple positive classes are provided for training along with unlabeled data from a mixture of the positive classes and a negative class \cite{xu2017multi}.


\paragraph{Multiple annotators} Assuming we have $m$ annotators, we also have potentially $m$ distinct mixing matrices \cite{raykar,ni2013understanding,perello2020recycling}. Therefore, $\tilde{Y}^n~=~\big\{\tilde{Y} \sim \texttt{Cat}\big(\boldsymbol{T}_j\boldsymbol{y}\big)\big\}_{j=1}^n$, where $\boldsymbol{T}_j$ denotes the $j^{th}$ annotator's mixing matrix. 


\begin{align*}
    \mathcal{Y}_k&~\to~\mathcal{Y}^{n}_{k}\\
    \Big\{\boldsymbol{T}&~\in~\mathbb{R}^{k\times k}\Big\}_{j=1}^n
\end{align*}
where,
\begin{equation}
    \label{eq:multiannotator_labelspace}
    \mathcal{Y}_{k}^n ~=~ \big\{\boldsymbol{y}~|~\boldsymbol{y}\in\{0,~1\}^{k\times n},~\boldsymbol{y}^{\top}\boldsymbol{1}=\boldsymbol{1}\big\}
\end{equation}
where the equality should be understood as elementwise.

\subsection{Aggregate WSL Settings}
Aggregate settings have comparatively received much less attention in the literature. Here, we present multiple instance learning and learning from label proportions as the main representatives.
\paragraph{Multiple Instance Learning} Multiple instance learning (MIL) is usually considered in the binary classification case. The label provided for a bag of samples is an indicator of the presence of the positive class. In other words, is there at least one positive instance in the set? It was first introduced in \cite{dietterich1997solving} with the motivation of drug activity prediction.
\begin{equation*}
    g(y_1, \dots, y_n)=max\big(y_{[1..n]}\big)
\end{equation*}
In \cite{scott2005generalized} the authors extend MIL, from requiring at least one positive instance in a bag, to requiring $r$.
\begin{equation*}
    g(y_1, \dots, y_m)~=~\boldsymbol{1}\Big\{\sum_{i=1}^{m} y_{i} \geq r\Big\}
\end{equation*}

\paragraph{Learning from Label Proportions (LLP)}
In LLP, the aggregation function is the count function (or proportion) for each of the classes present in a bag of samples. LLP was introduced in \cite{musicant2007supervised} with the motivation of learning with mass spectrometry data.
\begin{equation*}
    g(y_1, \dots, y_m)=\Bigg[\sum_{i=1}^{m} y_{i,0},~\cdots,~\sum_{i=1}^{m} y_{i,c}\Bigg]
\end{equation*}
Interestingly, PU Learning was presented as being a non-aggregate setting, but under certain conditions it can also be seen as a case of learning from label proportions. In PU Learning, we are provided with a set of positive data and a set of unlabelled data. If we know the portion of positives and negatives in the unlabelled set, we can view this as an instance of LLP with two bags.

\subsection{Towards a Unified Formulation}

For non-aggregate settings, we can now see how to find a common formulation for weakly supervised settings. Starting with the label space corresponding to multiple annotators in Eq. \ref{eq:multiannotator_labelspace}, we can extend it to allow annotators to provide an empty set for a sample (e.g., when they find an instance difficult to annotate):
\begin{equation}
    \label{eq:multiannotator_w_unlabelled_labelspace}
    \mathcal{Y}_{k+}^n ~=~ \big\{\boldsymbol{y}~|~\boldsymbol{y}\in\{0,~1\}^{k\times n},~\boldsymbol{y}^{\top}\boldsymbol{1}\in\{0,1\}\big\}
\end{equation}
and then it could be extended to allow annotators to provide an annotation set with more than one candidate,
\begin{equation}
    \label{eq:multiannotator_multlabel_labelspace}
    \mathcal{Y}_{m,k+}^n ~=~ \big\{\boldsymbol{y}~|~\boldsymbol{y}\in\{0,~1\}^{k\times n},~\boldsymbol{0}\leq\boldsymbol{y}^{\top}\boldsymbol{1}\leq \boldsymbol{m}\leq \boldsymbol{k}\big\}
\end{equation}
With regards to describing the annotation process for a dataset, we could model the weakening process through its dependence on the instance and the true class. 

Aggregate settings can also be extended to the case of having access to unsupervised data or multiple annotators. Instance-dependence has a different meaning in this case though. While previously it had to do with whether the weakening function was uniform across all instances, in this case it has to do with bag creation and intra-bag similarities \cite{zhou2009multi,carbonneau2018multiple,scott2020learning}

\section{Conclusion \& Future Work}
\label{section:conclusion}
We have presented a framework with dimensions that can help in navigating the weak supervision field, but that can also help in understanding exploiting the flexibility of the annotation process. 

However, this is nothing but a first step towards categorizing weak supervision. In future work we wish to extend this work in three directions. First, we aim to complement our framework with corresponding algorithms where a practitioner can turn to after identifying the characteristics of their problem. This would also allow for understanding the implications that certain choices on the annotation process have on available algorithms, their theoretical guarantees and practical considerations. Second, we want to expand this work into a transparent process for documenting the annotation of a dataset. This very much aligns with the proposal of accompanying a dataset with a datasheet that documents its motivation, composition, collection process and recommended uses in \textit{Datasheets for Datasets} \cite{gebru2021datasheets}. Third, we want to strengthen the framework itself. Non-aggregate weak supervision settings have been more widely studied and hence the difference in weight they have received in this paper. As seen, they can be unified through Eq.~\ref{eq:noise_categorigal} and the mixing matrix which can be used to reverse the noise process and make learning unbiased~\cite{van2015learning}. An important aspect in these settings is whether the mixing matrix is known~\cite{Bacaicoa-Barber21} or whether it has to be estimated~\cite{perello2020recycling}. Also, the unified formulation is not only a matter of aesthetics, but more importantly allows for the transferability of methods and theory. In the case of aggregate WSL settings, while we have Eq.~\ref{eq:aggregate_function} showing how aggregation is abstracted away, it does not improve our theoretical understanding, or allow for algorithms to be applied across settings yet. 


\bibliographystyle{IEEEtran}
{\small\bibliography{IEEEabrv,bibliography}}

\end{document}